\documentclass{article}
\PassOptionsToPackage{numbers,compress}{natbib}
\PassOptionsToPackage{table}{xcolor}

\usepackage[preprint]{neurips_2026}

\usepackage{wrapfig}
\usepackage[utf8]{inputenc}
\usepackage[T1]{fontenc}
\usepackage{hyperref}
\usepackage{url}
\usepackage{booktabs}
\usepackage{amsfonts}
\usepackage{nicefrac}
\usepackage{microtype}
\usepackage{xcolor}
\usepackage{float}
\usepackage{amsmath}
\usepackage{makecell}
\usepackage{enumitem}
\usepackage{arydshln}
\usepackage{graphicx}
\usepackage{multirow}

\usepackage{booktabs}
\usepackage{tabularx}
\usepackage{array}
\usepackage{xcolor}
\usepackage{hyperref}
\PassOptionsToPackage{table}{xcolor}
\usepackage{xcolor}

\newcommand{\best}[1]{\textbf{#1}}
\newcommand{\second}[1]{\underline{#1}}

\author{%
\raisebox{0.4em}[0pt][0pt]{%
\begin{tabular}{c}
\href{https://linghuiishen.github.io/BadWorld/}{\texttt{https://linghuiishen.github.io/BadWorld/}}\\[0.4em]
Linghui Shen \quad Mingyue Cui \quad Xingyi Yang\thanks{Corresponding author.} \\
The Hong Kong Polytechnic University \\
\texttt{\{ling-hui.shen, ming-yue.cui\}@connect.polyu.hk, xingyi.yang@polyu.edu.hk}
\end{tabular}%
}%
}
\begin{document}

\setlength{\abovedisplayskip}{2pt}      
\setlength{\belowdisplayskip}{2pt}      
\setlength{\abovedisplayshortskip}{0pt} 
\setlength{\belowdisplayshortskip}{3pt} 

\title{BadWorld: Adversarial Attacks on World Models}

\maketitle

\begin{abstract}

Visual world models (VWMs) synthesize interactive, action-conditioned rollouts from a single context image. However, it remains an open question how robust these models are to adversarial perturbations. Standard adversarial attacks fail to assess this vulnerability because attackers lack ground-truth future videos and cannot predict subsequent user controls. We introduce \textsc{BadWorld}, a label-free adversarial framework tailored for autoregressive VWMs that systematically overcomes both constraints. First, to bypass the need for future supervision, we propose a self-supervised \textit{velocity attack} that directly disrupts the early denoising dynamics of the model. Second, to ensure the attack generalizes across unpredictable user actions, we formulate a \textit{trajectory-adaptive bi-level optimization} that actively mines hard control sequences to forge control-agnostic perturbations. Evaluated on representative VWMs with continuous and discrete controls, \textsc{BadWorld} exposes severe structural fragility. Visually indistinguishable adversarial images reliably trigger catastrophic degradation in future rollouts, leading to incomplete denoising, structural collapse, and control inconsistency. These findings reveal critical risks for deploying VWMs in safety-critical systems while highlighting a practical mechanism for privacy protection.

\end{abstract}

\section{Introduction}

Visual world models are moving from passive video generators to interactive simulators. Given a single context image and a sequence of user-defined actions, they can synthesize action-conditioned future videos~\cite{https://doi.org/10.5281/zenodo.1207631, planet, dreamer}. This ability makes them useful for interactive games~\cite{genie, matrixgame2, yume, matrixgame3}, robotics~\cite{embodiedai_survey, RT-1, navigationworld, OpenX-Embodiment}, and autonomous driving~\cite{nuScenes, autodriving_survey, GAIA-1, DrivingWorld, drivetofuture, Epona}. As these models generate increasingly coherent rollouts, a common belief has emerged: visual world models may have implicitly learned physical and geometric rules of the world~\cite{phyicalbench, phyquestion, PhysGen, physical_law}.

This progress raises a fundamental robustness question: \emph{are the learned dynamics of current world models stable under small input perturbations?} This question is especially important for safety-critical applications, where fragile rollouts may undermine simulation, prediction, or planning. In this work, we use adversarial perturbations as a stress test for this form of temporal robustness.

However, existing adversarial attacks on generative models do not directly fit world models. Prior work mainly targets text-to-image personalization~\cite{caat, diffusionguard, disruptingdiffusion, antidreambooth, advdm, simac, glaze}, image-to-image editing~\cite{photoguard, facelock, decontext}, and conventional image-to-video or video generation pipelines~\cite{backdoorvideo, prime, dragI2Vattack, i2vguard, anti-i2v, vidfreeze, immune2v}. These attacks are usually designed for fixed generation conditions, reference outputs, or localized spatial-temporal degradation. In contrast, autoregressive world models are interactive, history-dependent, and control-conditioned~\cite{autoregressive_video, Astra, matrixgame2, genie, matrixgame3, yume}, which leads to two central challenges.
\vspace{-1.5mm}
\begin{itemize}[
    label=\raisebox{0.15ex}{\scriptsize$\triangleright$},
    leftmargin=1.5em,
    itemsep=0.25em,
    topsep=0.25em
]
    \item \textbf{\textsc{C1}: Missing future supervision.}
    A world-model adversary only observes a single context image, without paired future videos, ground-truth trajectories, or predefined correct rollouts. Therefore, reference-based adversarial losses are not directly applicable.

    \item \textbf{\textsc{C2}: Unknown future controls.}
    World-model rollouts depend on future camera paths, navigation commands, or discrete user actions. An attack optimized for one fixed trajectory may fail when the control signal changes.
\end{itemize}
\vspace{-1.5mm}
To address these challenges, we propose \textbf{BadWorld}, a label-free adversarial framework for autoregressive world models. Given a pretrained world model and a single clean context image, BadWorld learns an imperceptible perturbation that drives future rollouts into out-of-distribution behaviors, without requiring paired future videos or knowledge of the user's future actions. Specifically, BadWorld consists of two technical components.
\vspace{-1.5mm}
\begin{itemize}[
    label=\raisebox{0.15ex}{\scriptsize$\triangleright$},
    leftmargin=1.5em,
    itemsep=0.25em,
    topsep=0.25em
]
    \item \textbf{\textsc{S1}: Self-supervised velocity attack.}
    To address \textbf{\textsc{C1}}, we attack the model's predicted velocity space instead of comparing outputs with unavailable ground truth. We use the model's own denoising dynamics as supervision, together with an early-denoising approximation and a simple context-based history proxy. This yields a label-free attack that requires neither future videos nor action annotations.

    \item \textbf{\textsc{S2}: Trajectory-adaptive optimization.}
    To address \textbf{\textsc{C2}}, we formulate the attack as a bi-level optimization problem. The outer loop searches for hard trajectories under which the current perturbation is least effective, while the inner loop updates the perturbation against these trajectories. This produces a more control-agnostic adversarial image that is harder to bypass by simply changing the action sequence.
\end{itemize}
\vspace{-1.5mm}
We evaluate BadWorld on representative autoregressive world models with both continuous camera control and discrete action control. Our results show that current world models are surprisingly fragile: adversarial images remain visually close to clean inputs, yet the generated rollouts can suffer from incomplete denoising, structural collapse, semantic drift, and loss of control consistency. Across models and metrics, Velocity-Min consistently produces strong degradation, while trajectory-adaptive optimization further improves robustness across difficult trajectories.

Our findings reveal a robustness risk for deploying world models in safety-critical systems. They also suggest a practical privacy application: imperceptible perturbations can help protect images from unauthorized interactive generation.

Our contributions are summarized as follows: (1) We formalize adversarial attacks on autoregressive world models. (2) We propose BadWorld, a self-supervised attack framework that manipulates the model's predicted velocity and introduces four label-free objectives. (3) We introduce trajectory-adaptive bi-level optimization to produce perturbations that remain effective across diverse future controls. (4) We show that representative interactive world models are highly fragile; imperceptible perturbations can severely corrupt future rollouts, with implications for both safety and privacy.

\vspace{-2mm}
\section{Related Work}
\vspace{-1mm}
\subsection{Controllable Visual World Models}
\vspace{-1mm}

Recent advancements in diffusion and flow-matching frameworks \cite{diffusion, flowmatching, selfforcing, diffusionforcing, causalforcing} have shifted video generation from passive synthesis toward interactive simulation \cite{svd, skyreelsv2, Lumiere, Wan2.1, hunyuanvideo, moviegen, opensora, video_survey}. Unlike standard generators, visual world models (VWMs) aim to internalize learned physics and transition dynamics, evolving in response to agent actions and temporal history. A prominent trend involves adapting pretrained models into autoregressive frameworks for viewpoint-aware rollouts. Within this landscape, control is typically implemented through three pathways: explicit continuous camera trajectories \cite{recammastercamera, Astra, worldmem}, discrete action tokens \cite{matrixgame2, oasis, hunyuanworld1.5, lingbot}, or implicit text descriptions \cite{yume}. We focus on these autoregressive denoising VWMs, selecting models from both continuous and discrete paradigms to investigate their adversarial robustness. This assesses whether these interactive environments reliably maintain consistency and control adherence, which is essential for safety-critical deployment.

\vspace{-3mm}
\subsection{Adversarial Attacks on Generative Models}

\vspace{-2mm}
Adversarial research in generative modeling typically focuses on two objectives: assessing structural vulnerabilities and enhancing privacy protection. Early work predominantly targeted image generation, employing imperceptible pixel-level perturbations to disrupt generation process.
Studies on Text-to-Image (T2I) personalization employ training data poisoning to prevent unauthorized model fine-tuning \cite{caat, diffusionguard, disruptingdiffusion, antidreambooth, advdm, simac}, while Image-to-Image (I2I) attacks perturb input context images to trigger abnormal outputs for privacy protection \cite{photoguard, facelock, decontext}.
Building upon prior image-based work, adversarial paradigms have recently extended to video generation, where temporal dynamics introduce new attack surfaces. Specifically, identified threats include backdoor triggers in text prompts \cite{backdoorvideo}, jailbreaking techniques \cite{prime}, and adversarial trajectories in drag-based I2V systems \cite{dragI2Vattack}. Recent studies further investigate world-model attacks via physical-condition perturbation in driving scenes \cite{worldmodelsdreamwrong} or automated attack search for world agents \cite{wmattack}. 
However, they address condition-level or search-level attacks, while we study image-space attacks on video world models. Most relevant to our work are approaches that target the input images in I2V pipelines \cite{i2vguard, anti-i2v, vidfreeze, immune2v}.
While effective, these frameworks are primarily evaluated on standard diffusion-based backbones like SVD \cite{svd}, CogVideoX \cite{cogVideo}, or Wan2.1 \cite{Wan2.1}. A critical gap remains in evaluating the adversarial robustness of autoregressive video world models, particularly those conditioned on interactive signals like camera trajectories or action sequences. To bridge this gap, we propose the first adversarial attack framework tailored to unique architectures of autoregressive video world models.

\vspace{-4mm}
\section{Background}
\vspace{-2mm}
This section provides the necessary background for studying the robustness of video world models. We first review autoregressive video generation and its flow-matching formulation, which together establish the foundation for defining our adversarial objective.
\vspace{-3mm}
\subsection{Preliminary: Autoregressive Video Generation for World Models}
\vspace{-2mm}In the study, we focus on the autoregressive (AR) model for world generation. Unlike bidirectional generation, the AR approach decomposes a video sequence into $K$ segments (chunks) $\{z_1, \dots, z_K\}$. Given context frame $x \in \mathbb{R}^{H \times W \times 3}$, the model factorizes the joint distribution:
\vspace{-1mm}
\begin{equation}
p(z_{1:K} | x, \tau_{1:K}) = \prod_{i=1}^K p(z_i | z_{<i}, x, \tau_i)
\end{equation}
The control $\tau_i$ typically encodes camera motions or actions. Each segment $z_i$ is generated conditioned on context $x$, history $z_{<i}$ (via sliding window), control $\tau_i$, and an optional prompt $p$. 

Most recent AR world models instantiate each chunk generator with diffusion or flow-matching \cite{flowmatching} dynamics. For a target chunk $z_i$, Flow Matching constructs an interpolated latent between the data and Gaussian noise $\epsilon \sim \mathcal{N}(0,I)$, as that $z_t^i = (1-t)z_i + t\epsilon$.
The velocity network $v_\theta$ is trained to predict the transport direction $v^*
=
\frac{d z_t^i}{dt}
=
\epsilon-z_i$
by minimizing
\begin{equation}
\mathcal{L}(\theta)
=
\mathbb{E}_{i,t,\epsilon}
\left[
\left\|
v_\theta(z_t^i,t \mid z_{<i},x,\tau_i)
-
v^*
\right\|_2^2
\right].
\end{equation}
During inference, $z_i$ is generated via sequential self-rollout. While capturing temporal dependencies, this AR structure is sensitive to error accumulation, amplifying small perturbations over the rollout.

\vspace{-2mm}
\subsection{Problem Setup and Challenges}
\vspace{-2mm}

Given a pretrained autoregressive world model $\mathcal{G}_\theta$, we aim to construct an adversarial context image
\begin{equation}
x_{\mathrm{adv}} = x + \delta,
\qquad
\|\delta\|_{\infty} \le \eta,
\end{equation}
such that the generated rollout videos under the same control sequence becomes substantially different from the clean rollout. The ideal rollout-level objective is
\begin{equation}
    \delta^*
    =
    \arg\max_{\|\delta\|_{\infty}\le \epsilon}
    \mathcal{D}\!\left(
    \mathcal{G}_{\theta}(x+\delta,\tau_{1:K}),
    \mathcal{G}_{\theta}(x,\tau_{1:K})
    \right),
\label{global_obj}
\end{equation}
where $\mathcal{G}{\theta}(\cdot,\tau_{1:K})$ is the video generated under controls $\tau_{1:K}$, and $\mathcal{D}$ measures the difference between the adversarial and clean videos. Intuitively, the attack finds a small perturbation $\delta$ within budget $\eta$ that maximizes this difference. A large difference indicates OOD behavior, such as visual artifacts, temporal inconsistency, motion collapse, or semantic drift. However, Eq.~\ref{global_obj} presents two challenges.

\textbf{C1: Missing future supervision.}
The first challenge is that the attacker only observes the context image $x$ and has no ground-truth future video. Since there is no predefined ``correct'' rollout, the adversarial objective in Eq.~\ref{global_obj} cannot be directly implemented. The attack must therefore be label-free and self-supervised.

\textbf{C2: Unknown future controls.}
A second challenge comes from the interactive nature of world models. The generated rollout depends on future control signals, such as camera paths, navigation commands, or discrete user actions. However, these controls may be unknown at attack time and can vary across users. A perturbation optimized for one fixed trajectory may therefore fail when the control sequence changes. A robust attack should remain effective across diverse future controls.

These two challenges guide the design of our method. To address C1, we replace the unavailable rollout-level supervision with a self-supervised velocity-level attack objective that perturbs the model's denoising dynamics. To address C2, we introduce trajectory-adaptive optimization that actively searches for hard control trajectories and optimizes the perturbation against them.
\vspace{-4mm}

\section{Methodology}
\vspace{-3mm}

Following the two challenges identified, \textsc{BadWorld} consists of two main components. First, to address C1, we design a label-free velocity attack objective that uses the model's own denoising dynamics as supervision. Second, to address C2, we introduce trajectory-adaptive rollout optimization, which mines hard control trajectories and improves perturbation effectiveness across control signals.

\begin{figure}[t]
    \centering
    \includegraphics[width=1.0\linewidth]{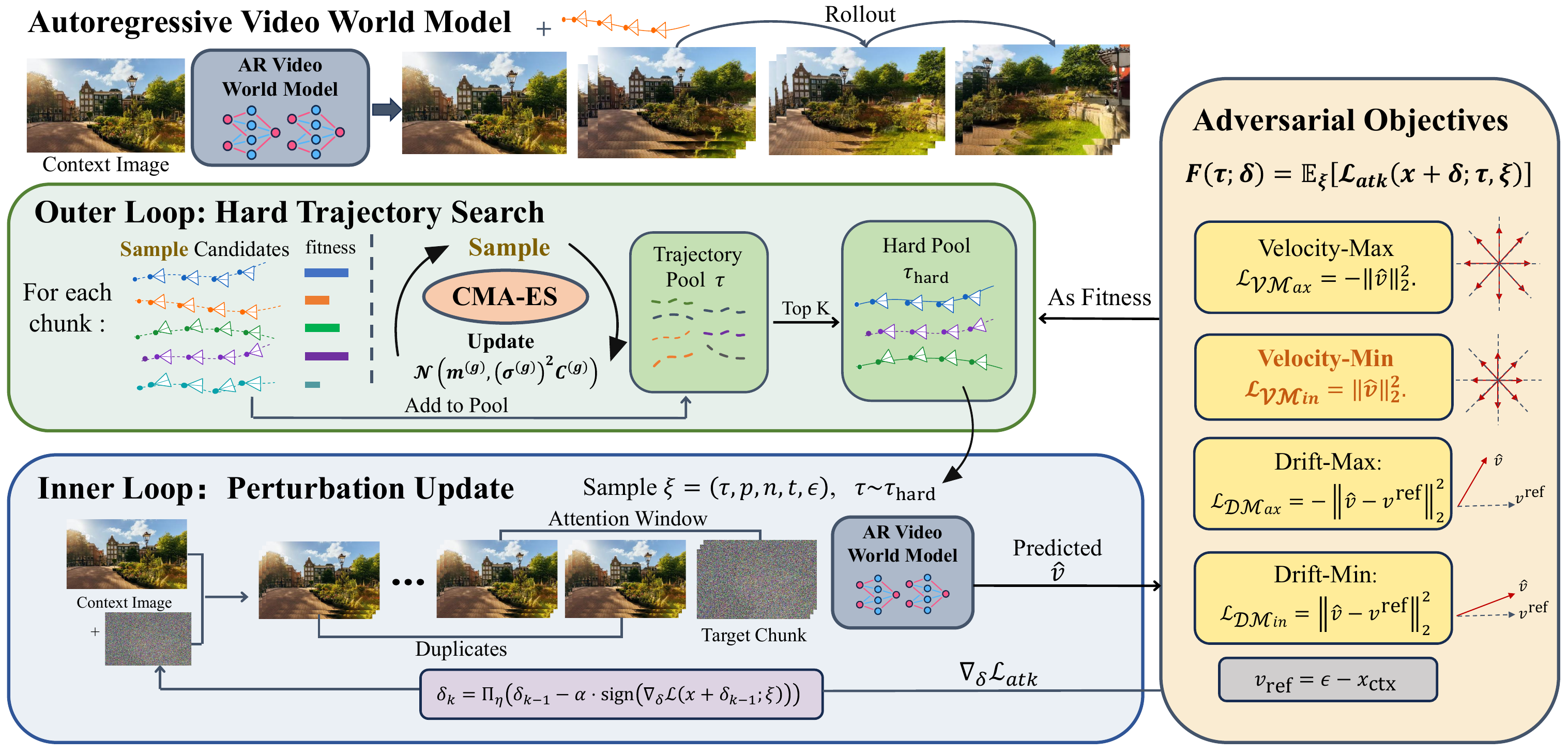}
    \vspace{-2mm}
    \caption{\textbf{BadWorld Pipeline.} An autoregressive video world model generates rollouts conditioned on context and camera (top). The outer loop employs CMA-ES for hard trajectory mining (middle), while the inner loop updates the adversarial perturbation under label-free settings (bottom). Four velocity-based objectives are provided (right), with Velocity-Min ($\mathcal{L}_{\mathrm{VMin}}$) serving as the default.} 
    \label{fig.model}
\vspace{-4mm}
\end{figure}
\vspace{-2mm}

\vspace{-2mm}
\subsection{Label-Free Velocity Attack Objective}
\vspace{-2mm}
\label{objs}
To address the absence of future supervision (C1), we avoid directly optimizing the final rollout discrepancy in Eq.~\ref{global_obj}. Since ground-truth future videos are unavailable in our setting, Eq.~\ref{global_obj} cannot be directly optimized. What's worse, such a rollout-level objective would require backpropagating through the entire sequence, which is computationally prohibitive.

Instead, we formulate the \emph{attack directly at the velocity level}. Since velocity determines the denoising direction, corrupted velocity predictions can disrupt local generation and further compound across autoregressive steps. We therefore optimize the perturbation $\delta$ to induce harmful velocity behavior.

Formally, we define a state tuple representing the generation context:
\begin{equation}
\xi = (\tau, p, n, t, \epsilon),
\end{equation}
where $\tau$ is a control signal, $p$ is an optional prompt, $n \in \{1,\dots,K\}$ denotes the target chunk, $t$ is a denoising timestep, and $\epsilon \sim \mathcal{N}(0,I)$ is Gaussian noise. Given the adversarial context $x+\delta$, the model predicts a velocity
\vspace{-1mm}
\begin{equation}
\hat{v}_{\delta}
=
v_{\theta}(\epsilon,t \mid h_n^{\delta}, \tau, p),
\label{eq:pred_velocity}
\end{equation}
where $h_n^{\delta}$ denotes the history condition used for the $n$-th autoregressive chunk. Using this notation, we define the general attack objective to find the optimal perturbation $\delta^*$ that minimizes an adversarial loss:
\vspace{-1mm}
\begin{equation}
\delta^* = \arg\min_{\|\delta\|_{\infty}\le \eta} \mathbb{E}_{\xi} \left[ \mathcal{L}_{\mathrm{atk}}(x+\delta;\xi) \right].
\label{eq:atk_objective}
\end{equation}

Here, $\mathcal{L}_{\mathrm{atk}}(x+\delta;\xi)$ specifies the harmful velocity property induced by the adversarial context. In practice, we estimate the expectation by random sampling and update $\delta$ using projected gradient descent~(PGD)\cite{pgd}. Specifically, we instantiate $\mathcal{L}_{\mathrm{atk}}$ through two types of self-supervised objectives.


\textbf{Velocity magnitude objectives.}
The first type of objective attack the \emph{magnitude} of the predicted velocity:
\begin{equation}
\mathcal{L}_{\mathrm{VMax}}(x+\delta;\xi)
=
-\|\hat{v}_{\delta}\|_2^2,
\qquad
\mathcal{L}_{\mathrm{VMin}}(x+\delta;\xi)
=
\|\hat{v}_{\delta}\|_2^2.
\label{eq:vmax_vmin}
\end{equation}
By maximizing this magnitude, Velocity-Max ($\mathcal{L}_{\mathrm{VMax}}$) forces aggressive denoising, which injects unnatural contrast and severe distortions into the chunk. Conversely, Velocity-Min ($\mathcal{L}_{\mathrm{VMin}}$) suppresses the update norm, causing the model to retain initial noise and yield incomplete structures. Because of the autoregressive nature of the model, these localized denoising failures compound, leading to severe degradation over long-horizon rollouts.

\textbf{Velocity direction objectives.}
The second type of objective attacks the \emph{velocity direction}. We define a context-anchored reference velocity as $v^{\mathrm{ref}} = \epsilon - x_{\mathrm{ctx}}$, where $x_{\mathrm{ctx}}$ is the encoded original image. This reference represents \emph{completely static video} that simply preserves the initial frame. Using this reference velocity, we define:
\begin{equation}
\mathcal{L}_{\mathrm{DMax}}(x+\delta;\xi)
=
-\|\hat{v}_{\delta}-v^{\mathrm{ref}}\|_2^2,
\qquad
\mathcal{L}_{\mathrm{DMin}}(x+\delta;\xi)
=
\|\hat{v}_{\delta}-v^{\mathrm{ref}}\|_2^2. \label{eq:dmin}
\end{equation}
Intuitively, Drift-Max ($\mathcal{L}_{\mathrm{DMax}}$) pushes the predicted velocity away from this static baseline video, encouraging semantic drift and temporal inconsistency. Conversely, Drift-Min ($\mathcal{L}_{\mathrm{DMin}}$) pulls the velocity toward the static reference, suppressing scene evolution and inducing motion collapse.

\textbf{Label-free realization.}
\label{label-free design}
Evaluating these objectives requires querying the predicted velocity $\hat{v}_{\delta}$. In standard flow-matching, computing the interpolated latent $z_t^n = (1-t)z_n + t\epsilon$ and the target velocity $v^*=\epsilon-z_n$ both require the future ground-truth chunk $z_n$. Since $z_n$ is unavailable in our label-free setting, we introduce two practical approximations.

\textit{Early-denoising approximation.}
When the timestep $t$ approaches $1$, the interpolated latent is overwhelmingly dominated by Gaussian noise:
\begin{equation}
z_t^n = (1-t)z_n + t\epsilon \approx \epsilon.
\end{equation}
By restricting queries to early denoising timesteps, we can input $\epsilon$ directly, entirely bypassing $z_n$. Attacking these early stages remains highly effective: early disruptions alter the global structure, seeding fundamental errors that drastically amplify throughout the autoregressive rollout.

\textit{Context-based history proxy.}
Autoregressive generation also demands a historical condition $z_{<n}$. Because the true future rollout is unavailable, we construct a differentiable proxy by repeating the encoded adversarial context:
\begin{equation}
h_n^{\delta}
=
\mathrm{Repeat}(E(x+\delta),L_n),
\label{eq:history_proxy}
\end{equation}
where $L_n$ is the required history length for the $n$-th chunk. This proxy provides a reliable pathway for the perturbation $\delta$ to influence the velocity prediction. Combined with the early-denoising approximation, this establishes the fully label-free velocity query (Eq.~\ref{eq:pred_velocity}), allowing us to optimize the objectives (Eq.~\ref{eq:vmax_vmin}--Eq.~\ref{eq:dmin}) without ground-truth future videos or explicit annotations.
\vspace{-3mm}

\vspace{-2mm}
\subsection{Trajectory-Adaptive Bi-Level Optimization}
\label{bi-attack}
\vspace{-3.5mm}

Although the self-supervised objectives in Sec.~\ref{objs} address \textbf{C1}, interactive world models remain sensitive to control signals $\tau$. Perturbations optimized for a single action sequence may overfit to its motion pattern and fail under unpredictable controls (\textbf{C2}). To ensure generalization across control sequences, we propose \textbf{trajectory-adaptive bi-level optimization}. The core intuition is that an attack that remains effective against hard trajectories is more likely to generalize to diverse user controls.

\vspace{-1mm}

\textbf{Min-Max Formulation.} Let $c_t$ be the camera pose at frame $t$, and $\tau=\{c_1,\dots,c_T\}$ denote a continuous camera trajectory over $T$ target frames. We frame our trajectory-adaptive attack as a min-max optimization game:
\begin{equation}
    \min_{\|\delta\|_\infty \le \eta}
    \;
    \max_{\tau \in \mathcal{T}}
    \;
    F(\tau;\delta),
    \qquad
    F(\tau;\delta)
    =
    \mathbb{E}_{\xi \sim \Omega(\tau)}
    \left[
    \mathcal{L}_{\mathrm{atk}}(x+\delta;\xi)
    \right],
    \label{eq:bilevel_main}
\end{equation}
where $\mathcal{T}$ is the feasible trajectory space, and $\Omega(\tau)$ denotes the distribution of label-free query states whose control component is fixed to the candidate trajectory $\tau$. The inner minimization updates the image perturbation $\delta$ to minimize the velocity surrogate under the current hard trajectories, while the outer maximization identifies hard trajectories that currently resist the attack and yield the highest expected loss $F(\tau;\delta)$.

\begin{figure}[t]
    \centering
    \includegraphics[width=0.98\linewidth, trim=0.1cm 0cm 0cm 0cm, clip]{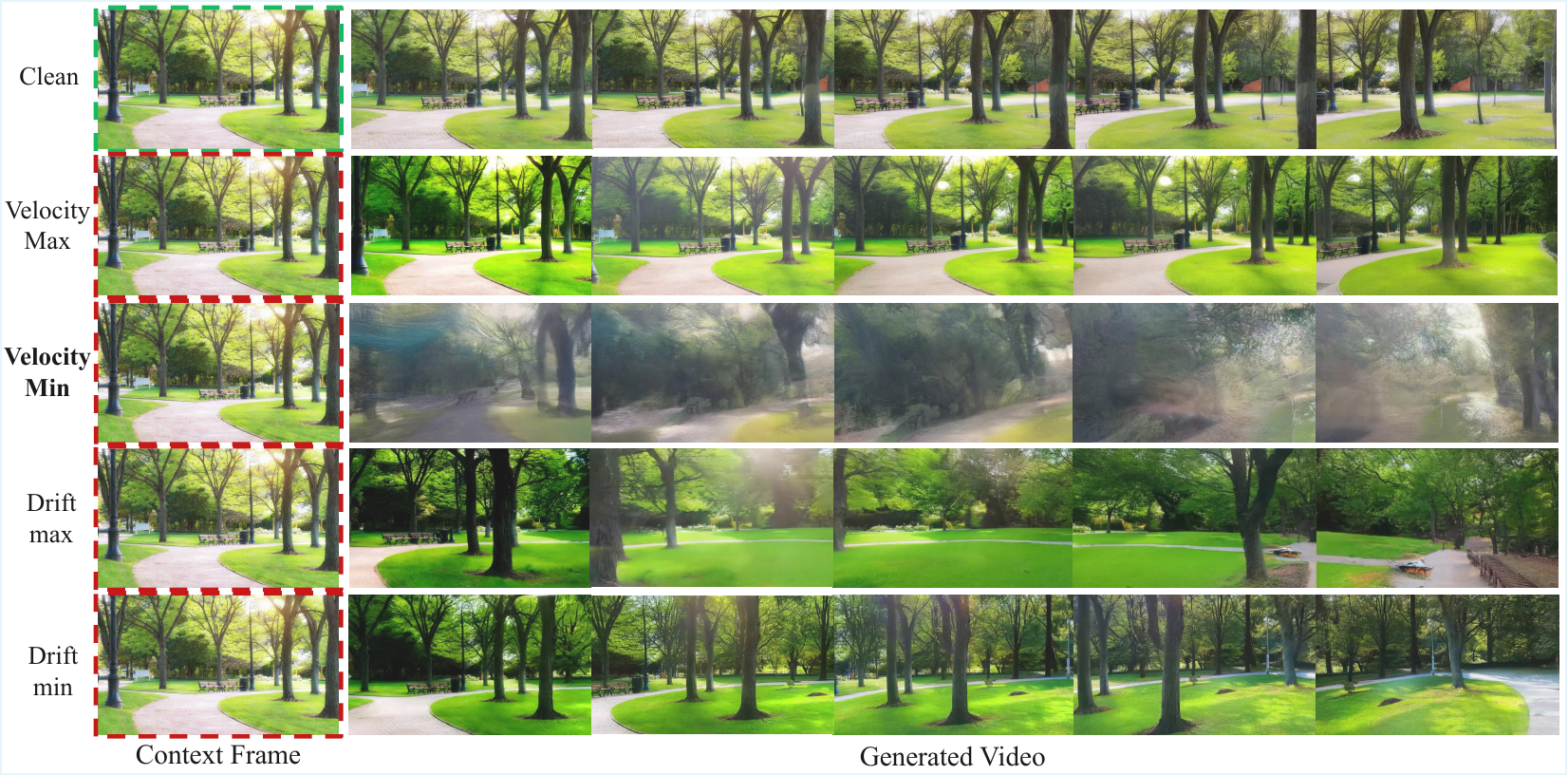}
    \vspace{-3mm}
    \caption{Qualitative comparison on Astra. Velocity-Max and Drift objectives induce color distortions. Velocity-Min causes severe geometric collapse.}
    \label{fig:astra_demo}
\vspace{-6mm}
\end{figure}

To parameterize the search space $\mathcal{T}$ efficiently, we discretize the continuous control at each frame into a compact vector preserving the primary degrees of freedom:
\begin{equation}
    c_t=(\psi_t, f_t, s_t),
    \qquad
    \tau=[\psi_1,f_1,s_1,\dots,\psi_T,f_T,s_T],
    \label{eq:traj_param}
\end{equation}
where $\psi_t$ (yaw), $f_t$ (forward translation), and $s_t$ (lateral shift) define the viewpoint transformation. Further conversion details are provided in the Appendix.

\noindent\textbf{Outer loop: hard trajectory mining.}
The outer maximization in Eq.~\ref{eq:bilevel_main} aims to discover ``hard'' trajectories that resist the perturbation. However, optimizing this sequence is non-trivial; the autoregressive generation process is largely non-differentiable with respect to the control signals.

To search without gradients, we employ the Covariance Matrix Adaptation Evolution Strategy (CMA-ES)\cite{cma}. Unlike independent random sampling, CMA-ES adapts a multivariate Gaussian distribution to capture coherent camera motion patterns that effectively challenge the current perturbation.

Specifically, at each generation $g$, we maintain a search distribution over the trajectory vector $\tau$:
\begin{equation}
    q^{(g)}(\tau)
    =
    \mathcal{N}\!\left(m^{(g)},(\sigma^{(g)})^2C^{(g)}\right),
    \qquad
    \tau_k^{(g)} \sim q^{(g)}(\tau),\quad k=1,\dots,\lambda .
    \label{eq:cma_sampling}
\end{equation}

where $m^{(g)}$, $\sigma^{(g)}$, and $C^{(g)}$ denote the mean, step-size, and covariance matrix. Each sampled trajectory is projected onto $\mathcal{T}$ and scored by $F(\tau;\delta)$ (Eq.~\ref{eq:bilevel_main}). CMA-ES then ranks the candidates and updates its distribution:
The mean $m$ update shifts the search toward harder trajectories, while the covariance $C$ update increases the probability of sampling successful correlated directions. This is crucial for camera control, where hard cases often arise from coherent temporal motions rather than independent frame changes (details refer to Appendix). 

After the final CMA generation, we collect evaluated candidates and retain the top-$K$ trajectories:
\begin{equation}
    \mathcal{P}_{\mathrm{hard}}^{(n)}
    =
    \operatorname{TopK}_{\tau \in \mathcal{S}^{(n)}}
    F(\tau;\delta),
    \label{eq:hard_pool}
\end{equation}
where $\mathcal{S}^{(n)}$ is the set of trajectories evaluated for chunk $n$. We maintain a separate pool $\mathcal{P}_{\mathrm{hard}}^{(n)}$ for each chunk because trajectory difficulty varies across autoregressive steps.

\noindent\textbf{Inner loop: perturbation update.}
Given the hard trajectories identified by the outer loop, the inner loop updates the adversarial perturbation. At each PGD step, for a selected chunk $n$, we sample a trajectory $\tau$ from $\mathcal{P}_{\mathrm{hard}}^{(n)}$ and construct the query state $\xi_\tau$. The perturbation is updated by:
\begin{equation}
    \delta_{k+1} = \Pi_{\eta} \left( \delta_k - \alpha \operatorname{sign} \left( \nabla_{\delta} \mathcal{L}_{\mathrm{atk}}(x+\delta_k; \xi_\tau) \right) \right),
    \label{eq:inner_update}
\end{equation}
where $\Pi_{\eta}$ projects the perturbation back into the valid $\ell_\infty$ bounds.

We alternate between hard-trajectory mining and perturbation updates, producing an adversarial context image that induces unstable autoregressive rollouts across diverse controls.

\begin{figure}[t]
    \centering
    \includegraphics[width=0.98\linewidth]{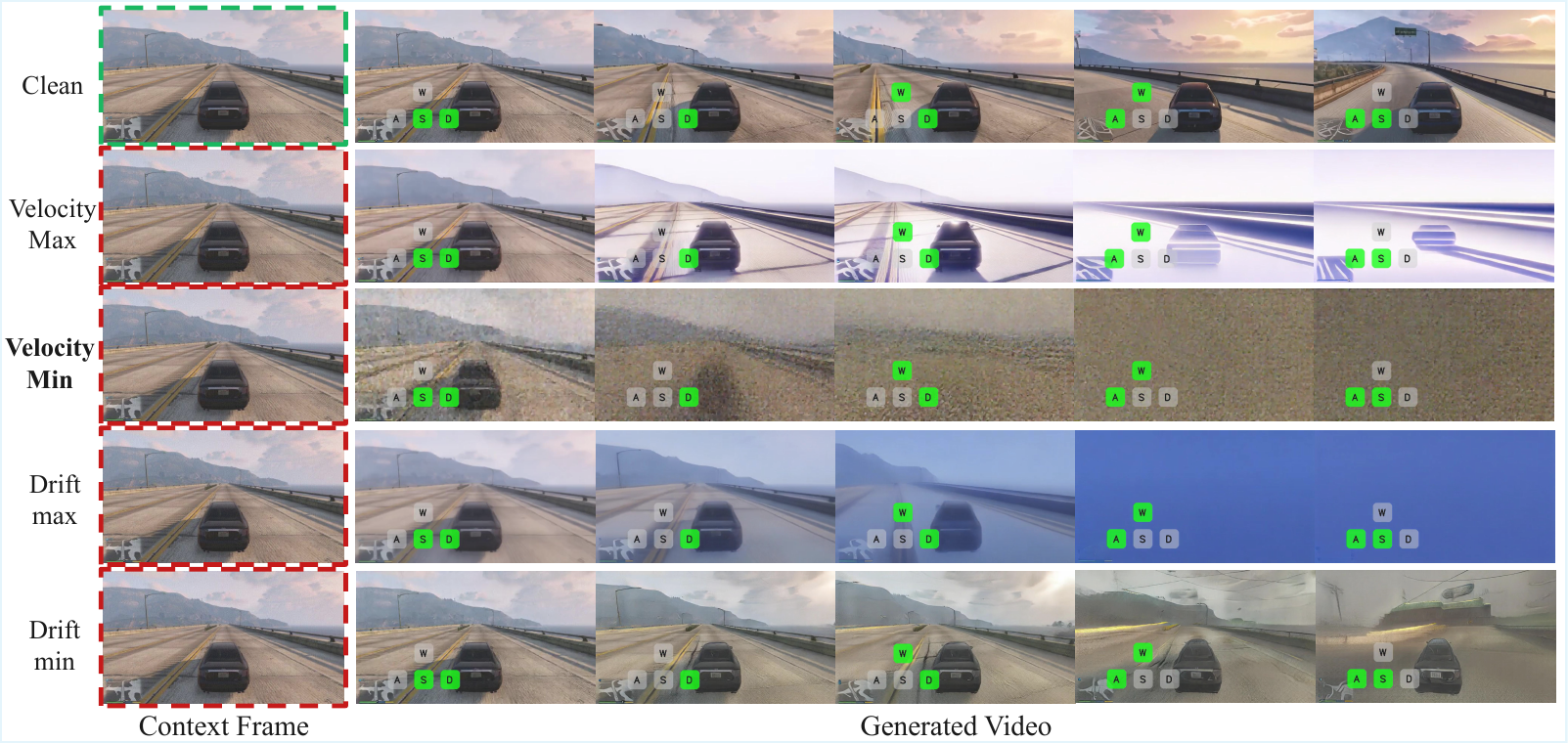}
    \vspace{-2mm}
    \caption{Qualitative comparison on Matrix-Game-2.0. Velocity-Max and Drift-Max inflate visual contrast. Drift-Min produces near-static videos. Velocity-Min causes complete structural collapse.}
    \label{fig:matrix_demo}
    \vspace{-3mm}
\end{figure}
\vspace{-4mm}
\begin{figure}[t]
    \centering
    \includegraphics[width=0.98\linewidth]{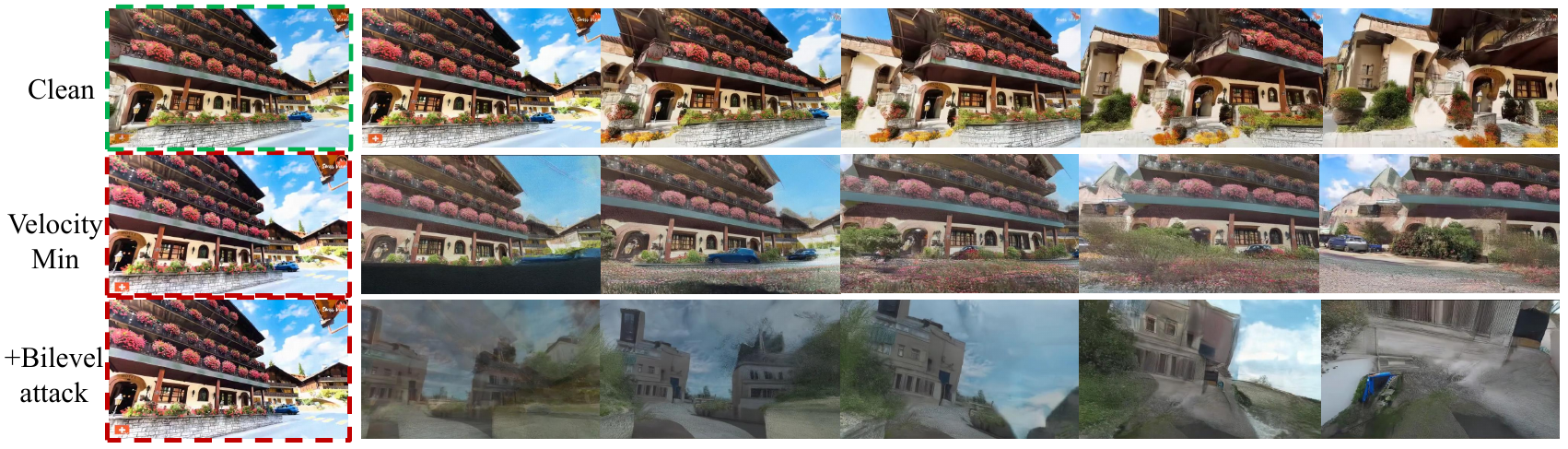}
    \vspace{-2mm}
    \caption{Bi-level Attack performance. }
    \label{fig.bilevel_demo}
\vspace{-7mm}
\end{figure}

\vspace{1mm}
\section{Experiments}
\vspace{-2mm}
\subsection{Experimental Setup}
\label{exp setup}
\vspace{-2mm}
\noindent\textbf{Models.}
We evaluate on two representative open-source autoregressive video world models:
\vspace{-2mm}
\begin{itemize}
\vspace{-1mm}
    \item \textbf{Astra} \cite{Astra}: Built on Wan2.1 \cite{Wan2.1}, Astra is fine-tuned for causal denoising and continuous camera pose conditioning, making it ideal for studying camera-driven adversarial behaviors.
\vspace{-2mm}
    \item \textbf{Matrix-Game 2.0} \cite{matrixgame2}: A lightweight, causal few-step generative model distilled from SkyReelsV2-I2V-1.3B \cite{skyreelsv2}. Utilizing discrete action signals, it provides a complementary platform to test robustness under categorical control.
\vspace{-1mm}
\end{itemize}

\noindent\textbf{Dataset Setup.} 
Since no established benchmark exists for adversarial attacks on video world models, we construct two 100-image datasets tailored for each model.
For Astra, we sample the first frame from natural landscape videos in the SpatialVID dataset \cite{spatialvid} and resized to $832 \times 480$.
For Matrix-Game 2.0, we extract \textit{Grand Theft Auto (GTA)} gameplay frames and resize to $640 \times 352$. To align with the training domain of Matrix-Game-2.0 and enhance clean generation quality, we apply Flux Kontext \cite{fluxkontext} to standardize primary object appearances. Specifically, vehicles are transformed into black sedans, and human subjects are edited to wear black clothing.

\noindent\textbf{Baselines.}
To the best of our knowledge, no prior work has studied adversarial attacks on autoregressive video generation for world models. 
Following I2VGuard~\cite{i2vguard}, we therefore adopt a noise baseline by adding random noise to the context frame, matching the mean and variance of the perturbations produced by our method.
 
\noindent\textbf{Evaluation Metrics.}
To assess adversarial impact, we employ VBench\cite{vbench}, VBench++\cite{vbench++}, CLIP \cite{clipi} and MEt3R\cite{met3r} to evaluate four dimensions.
First, we measure \textbf{contextual consistency} by accessing \textit{i2v subject} and \textit{i2v background}, which quantifies how well the generated video preserves identity and environmental details from the source image. We exclude i2v-subject metrics for Astra, as the dataset lacks explicit foreground subjects. Second, we evaluate \textbf{video quality} by reporting \textit{background consistency}, \textit{aesthetic quality}, and \textit{imaging quality} to capture visual degradation. Third, we evaluate \textbf{semantic preservation} by computing the \textit{CLIP-I} \cite{clipi} score between the first and last frames.
Fourth, we measure \textbf{geometry reliance} using \textit{MEt3R}~\cite{met3r}, which evaluates geometric consistency by reconstructing dense 3D geometry for each frame pair and measuring feature mismatch after cross-view projection. 

\noindent\textbf{Implementation Details.}
We execute all attacks on an A800 80GB GPU, simulating realistic label-free scenarios by restricting optimization to early denoising timesteps and duplicating context frames. During inference, we autoregressively generate three diverse videos per input using default model configurations. Comprehensive implementation details regarding model-specific conditionings and training hyperparameters are deferred to the Appendix.

\vspace{-4mm}
\subsection{Main Results on Different Attack Objectives}
\vspace{-2mm}
To compare the adversarial objectives introduced in Sec.~\ref{objs}, we attack 100 images per model under a noise budget of \(\eta = 0.05\), evaluating each image across three distinct camera or action sequences.
\vspace{-1mm}
\begin{table}[t]
\centering
\scriptsize 
\setlength{\tabcolsep}{3pt} 
\resizebox{0.85\textwidth}{!}{ 
\begin{tabular}{lcccccc}
\toprule
\textbf{Method} 
& \makecell{\textbf{I2V}\\ \textbf{Background} $\downarrow$}
& \makecell{\textbf{Background}\\ \textbf{Consistency} $\downarrow$}
& \makecell{\textbf{Aesthetic}\\ \textbf{Quality} $\downarrow$}
& \makecell{\textbf{Imaging}\\ \textbf{Quality} $\downarrow$}
& \textbf{CLIP-I$\downarrow$}
& \textbf{MEt3R$\uparrow$}\\
\midrule
Clean 
& 0.978 
& 0.915 
& 0.501 
& 0.690 
& 0.813
& 0.151\\
\hdashline
Random Noise
& 0.973
& 0.912
& 0.490
& 0.680
& 0.810
& 0.150\\
\rowcolor{yellow!15}
\multicolumn{7}{l}{\textbf{\textit{Self-supervised Attack Objectives}}}\\
Velocity-Max 
& 0.956 
& 0.891 
& 0.493 
& 0.657 
& 0.779
& 0.154\\
\textbf{Velocity-Min} 
& \best{0.930} 
& \best{0.845} 
& \best{0.405} 
& \best{0.513} 
& \best{0.714}
& \best{0.264}\\
Drift-Max 
& \second{0.938} 
& \second{0.866} 
& \second{0.466} 
& \second{0.564} 
& \second{0.750}
& \second{0.207}\\
Drift-Min 
& 0.968 
& 0.908 
& 0.475 
& 0.601 
& 0.800
& 0.166\\
\rowcolor{red!12}
\multicolumn{7}{l}{\textbf{\textit{Trajectory Adaptive Bi-level Attack}}}\\
VMin+Bilevel
& 0.925
& 0.842
& 0.396
& 0.524
& 0.712
& 0.265\\
\bottomrule
\end{tabular}
}
\vspace{1mm}
\caption{Comparison of attack objectives against Astra. }
\vspace{-5mm}
\label{tab:astra_main}
\end{table}
\begin{table}[t]
\centering
\small
\resizebox{\linewidth}{!}{
\setlength{\tabcolsep}{5pt}
\begin{tabular}{lccccccc}
\toprule
\textbf{Method} 
& \makecell{\textbf{I2V}\\ \textbf{Subject} $\downarrow$}
& \makecell{\textbf{I2V}\\ \textbf{Background} $\downarrow$}
& \makecell{\textbf{Background}\\ \textbf{Consistency} $\downarrow$}
& \makecell{\textbf{Aesthetic}\\ \textbf{Quality} $\downarrow$}
& \makecell{\textbf{Imaging}\\ \textbf{Quality} $\downarrow$}
& \makecell{\textbf{CLIP-I} $\downarrow$}
& \textbf{MEt3R $\uparrow$} \\
\midrule
Clean 
& 0.897 
& 0.897 
& 0.973 
& 0.512 
& 0.648 
& 0.796
& 0.098\\
\hdashline
Random Noise
& 0.878
& 0.895
& 0.974
& 0.522
& 0.647
& 0.785
& 0.101\\
\rowcolor{yellow!15}
\multicolumn{8}{l}{\textbf{\textit{Self-supervised Attack Objectives}}}\\
Velocity-Max 
& 0.788 
& 0.795 
& \second{0.964} 
& 0.443 
& 0.491 
& 0.631
& 0.119\\
\textbf{Velocity-Min} 
& \best{0.708} 
& \second{0.750} 
& \best{0.960} 
& \second{0.352} 
& \second{0.442} 
& \second{0.584}
& \best{0.204}\\
Drift-Max 
& \second{0.725} 
& \best{0.743} 
& 0.965 
& \best{0.340} 
& \best{0.364} 
& \best{0.537}
& \second{0.159}\\
Drift-Min
& 0.868 
& 0.868 
& 0.964 
& 0.478 
& 0.623 
& 0.631
& 0.101\\
\bottomrule
\end{tabular}
}
\caption{Comparison of attack objectives against Matrix-Game-2.0 (GTA).}
\vspace{-7mm}
\label{tab:matrix_main}
\end{table}

\noindent\textbf{Comparison on Astra.}
As shown in Tab. \ref{tab:astra_main} (best results are bolded, and second-best results are underlined), \textit{Velocity-Min} is the most effective objective, achieving the lowest scores across nearly all metrics. Compared to the clean baseline, it induces the sharpest decline in visual fidelity, reducing aesthetic quality from 0.501 to 0.405 and imaging quality from 0.690 to 0.513. This trend is further evidenced by MEt3R, where \textit{Velocity-Min} reaches $0.264$, representing a $74.8\%$ increase over the clean baseline. \textit{Drift-Max} follows as a strong runner-up, notably reducing imaging quality to $0.564$. Conversely, \textit{Velocity-Max} causes only moderate degradation, while \textit{Random Noise} and \textit{Drift-Min} have negligible impact, closely mirroring the clean baseline. For qualitative comparison, Fig.\ref{fig:astra_demo} shows that \textit{Velocity-Min} triggers incomplete denoising and grayish artifacts, whereas other objectives primarily alter color temperature.

\noindent\textbf{Comparison on Matrix-Game-2.0.} As detailed in Tab. \ref{tab:astra_main}, Matrix-Game-2.0 exhibits a broader vulnerability, where most objectives serve as effective attacks. \textit{Velocity-Min} remains the most potent, significantly reducing background consistency to $0.845$, imaging quality to $0.513$, and achieving the highest MEt3R of $0.264$. Other strategies like \textit{Drift-Max} and \textit{Velocity-Max} also demonstrate clear effectiveness by degrading imaging quality and CLIP-I. As visualized in Fig. \ref{fig:matrix_demo}, 
\textit{Velocity-Min} induces severe structural disintegration and noise, while \textit{Drift-Max} and \textit{Velocity-Max} primarily generate frames with unnaturally high color saturation. \textit{Drift-Min} tends to produce near-static videos that fail to capture the intended motion.

\noindent\textbf{Summary.} In summary, while Astra shows high selectivity with \textit{Velocity-Min} being the only truly effective objective, Matrix-Game-2.0 is susceptible to a wider range of perturbations. Across both benchmarks, \textit{Velocity-Min} consistently remains the strongest objective for degrading video quality.

We therefore adopt the strongest objective, \textit{Velocity-Min}, for all subsequent experiments. Furthermore, since baseline attacks on Matrix-Game-2.0 often trigger a near-complete collapse that masks performance nuances, we utilize the more challenging Astra benchmark to conduct ablation studies (\ref{ablations}) and validate the efficacy of our bi-level optimization framework (\ref{bilevel_exp}).

\vspace{-4mm}
\subsection{Ablation Studies}
\label{ablations}
\vspace{-2mm}
Following the experimental setup described in Sec. \ref{exp setup}, we conduct three ablation studies on Astra \cite{Astra} to evaluate the impact of different budgets and the effectiveness of our proposed designs.

\noindent\textbf{Ablation on attack budget.} We first examine the impact of the perturbation budget $\eta$ on attack performance. As shown in Tab. \ref{tab:ablation_budget}, increasing $\eta$ from 0.03 to 0.10 leads to a consistent decline in video quality metrics and a corresponding rise in MEt3R scores, indicating more effective disruption. However, larger budgets also make the adversarial perturbations more visually perceptible. To achieve a balance between attack potency and imperceptibility, we select $\eta = 0.05$ as our default setting.

\noindent\textbf{Ablation on diffusion timesteps.} 
We assess timestep selection by comparing our \textit{early-denoising} strategy (Sec.\ref{label-free design}) against uniform timestep sampling from $0$ to $1000$, denoted as ``-Timesteps Selection'' in Tab. \ref{tab:ablation_design}. Without this early focus, attack efficacy collapses: Aesthetic Quality rises to $0.514$ and MEt3R drops to $0.167$. Targeting initial structural formation is thus essential to effectively neutralize generative priors.

\noindent\textbf{Ablation on history simulation.} To verify the effectiveness of our \textit{context-based history proxy} (Sec. \ref{label-free design}), we compare it against a \textit{self-rollout} approach, denoted as ``+ Self-Rollout'' in Tab. \ref{tab:ablation_design}. Since performing rollouts at every attack step is computationally prohibitive, we use an interval-based scheme: every $50$ steps, we generate four videos from current adversarial context under distinct trajectories to update a history pool. During the subsequent interval, we sample from this pool as the history reference. This sophisticated approach yields negligible improvements, with metrics like MEt3R and CLIP-I showing slight improvement. Given the heavy rollout overhead, duplicating context frames is a more efficient and effective strategy.

\begin{table}[t]
\centering
\footnotesize
\setlength{\tabcolsep}{4pt}
\renewcommand{\arraystretch}{1.05}
\resizebox{\linewidth}{!}{
\begin{tabular}{lccccccc}
\toprule
\textbf{Budget} 
& \textbf{Step Size}
& \makecell{\textbf{I2V Back.} $\downarrow$}
& \makecell{\textbf{Back. Consis.} $\downarrow$}
& \makecell{\textbf{Aesthetic} $\downarrow$}
& \makecell{\textbf{Imaging Qual.} $\downarrow$}
& \textbf{CLIP-I $\downarrow$}
& \textbf{MEt3R $\uparrow$} \\
\midrule
$\eta = 0.03$ & 0.003 & 0.945 & 0.860 & 0.429 
& 0.563 & 0.710 & 0.272 \\
$\eta = 0.05$* & 0.005 & 0.930 & 0.845 & 0.405 
& 0.513 & 0.714 & 0.264 \\
$\eta = 0.10$ & 0.005 & \textbf{0.903} & \textbf{0.794} & \textbf{0.387} 
& \textbf{0.507} & \textbf{0.678} & \textbf{0.291} \\
\bottomrule
\end{tabular}
}
\caption{Performance under different attack budgets. * indicates the default setting.}
\vspace{-5mm}
\label{tab:ablation_budget}
\end{table}
\begin{table}[t]
\centering
\footnotesize
\setlength{\tabcolsep}{4pt}
\renewcommand{\arraystretch}{1.05}
\resizebox{\linewidth}{!}{
\begin{tabular}{lcccccc}
\toprule
\textbf{Method} 
& \makecell{\textbf{I2V Back.} $\downarrow$}
& \makecell{\textbf{Back. Consis.} $\downarrow$}
& \makecell{\textbf{Aesthetic} $\downarrow$}
& \makecell{\textbf{Imaging Qual.} $\downarrow$}
& \textbf{CLIP-I $\downarrow$}
& \textbf{MEt3R $\uparrow$}\\
\midrule
Velocity-Min 
& \textbf{0.930}
& 0.845
& \textbf{0.405}
& \textbf{0.513}
& 0.714
& 0.264\\
+Self-Rollout
& 0.932
& \textbf{0.843}
& 0.410
& 0.514
& \textbf{0.701}
& \textbf{0.271}\\
-Timesteps Selection
& 0.976
& 0.913
& 0.514
& 0.715
& 0.801
& 0.167\\
\bottomrule
\end{tabular}
}
\caption{Ablation on label-free designs.}
\vspace{-5mm}
\label{tab:ablation_design}
\end{table}
\begin{table}[H]
\centering
\small
\resizebox{\textwidth}{!}{
\begin{tabular}{lcccccc}
\toprule
\textbf{Method} 
& \makecell{\textbf{I2V Back.} $\downarrow$}
& \makecell{\textbf{Back. Consis.} $\downarrow$}
& \makecell{\textbf{Aesthetic} $\downarrow$}
& \makecell{\textbf{Imaging Qual.} $\downarrow$}
& \textbf{CLIP-I$\downarrow$}
& \textbf{MEt3R$\uparrow$}\\
\midrule
Velocity-Min 
& 0.935
& 0.845
& 0.440
& \textbf{0.556}
& 0.720
& 0.249\\
+ Bi-level Attack
& \textbf{0.931}
& \textbf{0.835}
& \textbf{0.425}
& \textbf{0.556}
& \textbf{0.711}
& \textbf{0.256}\\
\bottomrule
\end{tabular}
}
\caption{Hard Sample Performance Comparison.}
\vspace{-3mm}
\label{tab:hard_sample}
\end{table}

\vspace{-4mm}
\subsection{Performance of Trajectory-Adaptive Bi-level Optimization}
\label{bilevel_exp}
\vspace{-2mm}

Following the baseline protocol, we evaluate the bi-level attack on 100 Astra images across three trajectories each. As shown in Tab.~\ref{tab:astra_main}, trajectory-adaptive optimization further improves attack potency over the standard Velocity-Min objective, though the overall gains are moderate on the full benchmark. This is expected, as Velocity-Min is already highly effective for most samples, leaving limited room for further degradation.

To further assess robustness under more challenging cases, we target the 10 most resilient samples from the Velocity-Min baseline, evaluating each against seven distinct trajectories (Tab.~\ref{tab:hard_sample}). Our method consistently degrades all metrics, reducing Aesthetic Quality to 0.425. As illustrated in Fig.~\ref{fig.bilevel_demo}, bi-level optimization triggers significantly more pronounced noise and geometric distortion where the baseline struggles. These results confirm its superior generalizability in ensuring rollout collapse regardless of sample difficulty or camera path.
\vspace{-2mm}

\section{Conclusion}
\vspace{-2mm}
In this work, we present BadWorld, an adversarial framework designed for autoregressive visual world models (VWMs). To bypass the need for ground-truth future videos, we propose a self-supervised velocity attack that directly disrupts early denoising dynamics. To handle unpredictable future controls, we formulate a trajectory-adaptive bi-level optimization that mines hard control sequences to forge control-agnostic perturbations.
These findings expose critical safety risks for VWM deployment, yet provide a practical mechanism for privacy protection.

\bibliographystyle{plainnat}
\bibliography{references}

\newpage
\appendix

\begin{center}
    {\LARGE \bfseries Supplementary Material}\\[0.8em]
    {\Large \bfseries \textit{BadWorld: Adversarial Attack on World Models}}
\end{center}

In this supplement, we provide further details on BadWorld, including objective explanations, robustness evaluations, implementation specifics, and additional experimental results. Specifically, Sec.~\ref{sec:supp_velocity_magnitude} offers a deeper explanation of the velocity magnitude objective, Sec.~\ref{robustness} evaluates adversarial robustness against image preprocessing and black-box transferability across models, Sec.~\ref{app:implementation} details trajectory sampling algorithms and experimental hyperparameters, Sec.~\ref{sec:supp_extension_experiments} presents extension experiments on Matrix-Game-2 variants, and Sec.~\ref{sec:supp_qualitative_results} provides further qualitative visuals of our attack.
\vspace{-3mm}
\section{Explanation of the Velocity Magnitude Objective}
\label{sec:supp_velocity_magnitude}
\vspace{-3mm}
This section further explains the velocity magnitude objectives introduced in the main paper.

In flow-matching \cite{flowmatching} frameworks, the denoising network predicts a velocity field $\hat{v}_\theta$ specifying the direction and rate of latent-state change at each step. To isolate the effect of velocity magnitude, we perform a velocity scaling experiment during inference. Specifically, we modulate the predicted velocity by a positive scalar $s$, yielding $\hat{v}'_\theta = s \cdot \hat{v}_\theta$. This operation preserves the prediction direction and changes only its magnitude. The latent state is then updated via $z_{t-\Delta t} = \mathrm{SchedulerStep}(s \cdot \hat{v}_\theta, t, z_t)$ using the scaled velocity.

\begin{figure}[H]
    \centering
    \includegraphics[width=0.85\linewidth]{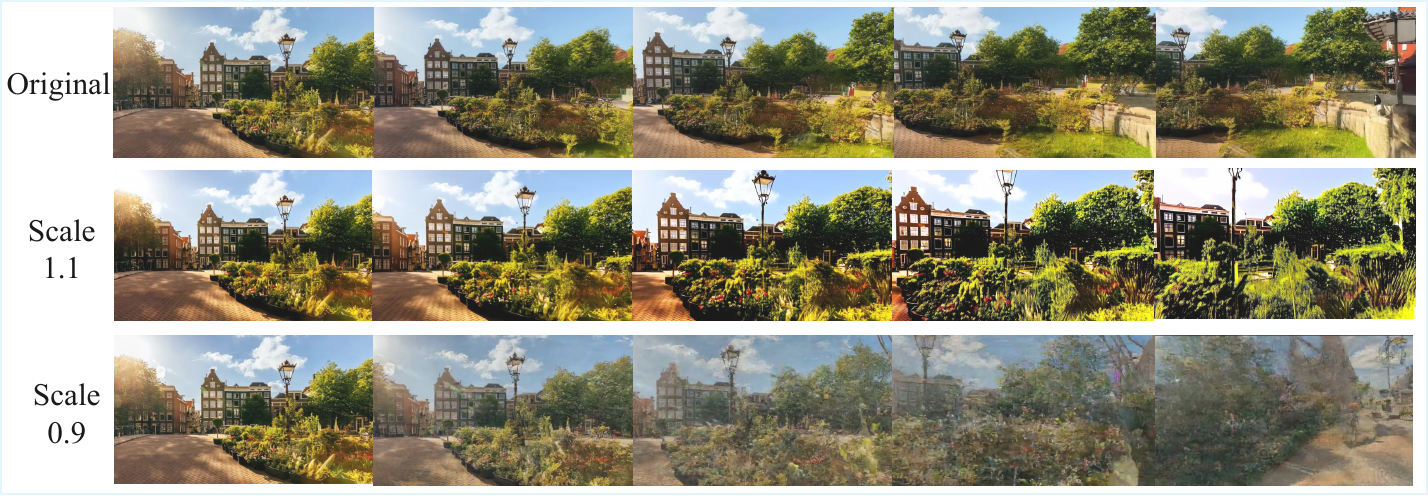}
    \vspace{-3mm}
    \caption{Scale the velocity magnitude for Astra.}
    \label{fig:astra_norm}
\end{figure}

\begin{figure}[H]
    \centering
    \includegraphics[width=0.85\linewidth]{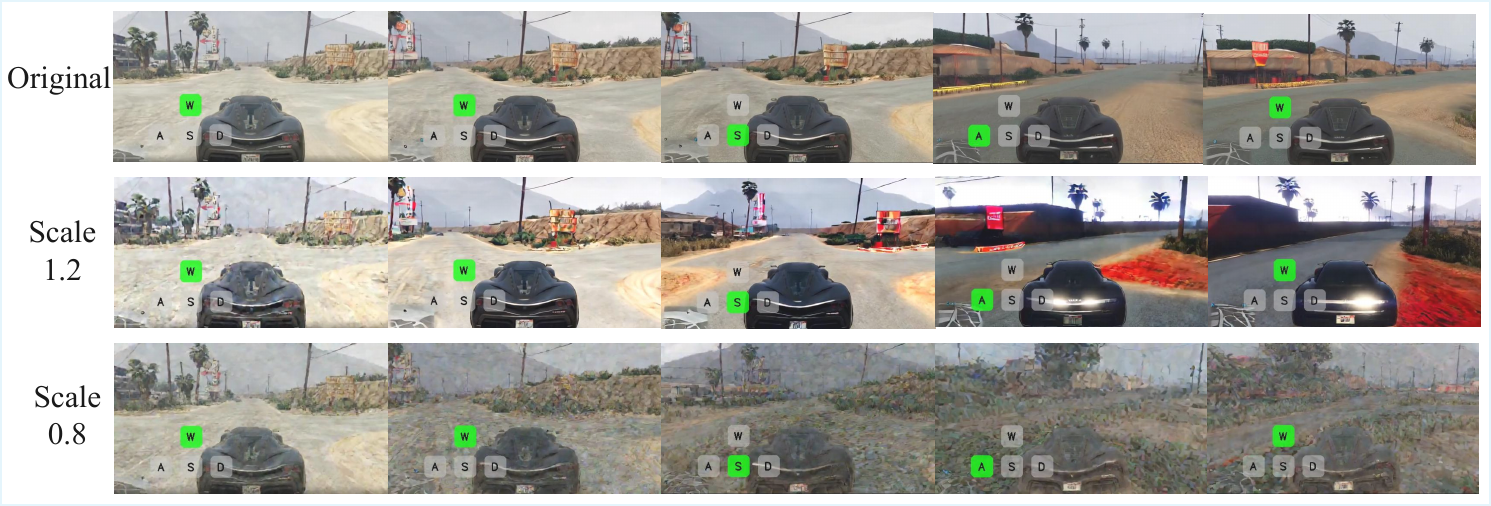}
    \vspace{-3mm}
    \caption{Scale the velocity magnitude for Matrix-Game-2.0. }
    \label{fig:matrix_norm}
\end{figure}

Empirical results in Fig.~\ref{fig:astra_norm} and~\ref{fig:matrix_norm} show that the magnitude of $\hat{v}_\theta$ directly determines visual quality. When velocity is under-scaled ($s < 1$), the update is too weak to drive the latent state toward the data manifold. This yields under-denoised outputs that appear gray, blurry, and structurally disordered. Conversely, over-scaling velocity ($s > 1$) pushes latents toward extreme values. While amplifying contrast, it often introduces overshooting artifacts such as extreme saturation and pixel distortion. These observations indicate that a precise velocity norm is vital for a stable denoising trajectory.

Building on these observations, we find that at equivalent scaling intensity, reducing velocity magnitude disrupts video coherence more severely than increasing it. For adversarial purposes, Velocity-Min proves superior to Velocity-Max in triggering rapid structural collapse. We hypothesize that this vulnerability stems from autoregressive world models: weakened updates prevent the model from reaching stable manifolds, causing subtle denoising errors to accumulate and compound through the temporal history window. This also motivates our choice of Velocity-Min as the final attack objective.
\vspace{-3mm}

\vspace{-5mm}

\newpage
\section{Robustness and Transferability}
\label{robustness}

Following the setup in Sec. \ref{imple_details}, we evaluate the robustness and transferability of our method. We randomly select 10 context images and generate videos across three distinct camera trajectories.

\noindent\textbf{Robustness to Image Preprocessing.}

We examine the resilience of our adversarial perturbations against common image preprocessing techniques, including Gaussian noise, JPEG compression, and Total Variation (TV) denoising. As shown in Tab. \ref{tab:img_preprocessing}, Velocity-Min remains highly effective under light Gaussian noise ($\sigma=0.5$), maintaining a strong MEt3R score of $0.255$ and significantly low imaging quality. While more aggressive preprocessings like JPEG and TV denoising partially mitigate the attack, our method still induces measurable degradation compared to the clean baseline. These results demonstrate that our perturbations are not easily neutralized by standard image-level filters, confirming their practical robustness.

\begin{table}[H]
\centering
\small
\setlength{\tabcolsep}{6pt}
\renewcommand{\arraystretch}{1.15}
\begin{tabular}{lcccccc}
\toprule
\textbf{Method} 
& \makecell{\textbf{I2V}\\ \textbf{Background} $\downarrow$}
& \makecell{\textbf{Background}\\ \textbf{Consistency} $\downarrow$}
& \makecell{\textbf{Aesthetic}\\ \textbf{Quality} $\downarrow$}
& \makecell{\textbf{Imaging}\\ \textbf{Quality} $\downarrow$} 
& \textbf{CLIP-I$\downarrow$}
& \textbf{MEt3R$\uparrow$}\\
\midrule
Clean 
& 0.978 
& 0.915 
& 0.501 
& 0.690 
& 0.813
& 0.151\\
Velocity-Min
& 0.922
& 0.816
& 0.405
& 0.506
& 0.701
& 0.266\\
\hdashline
+gaussian$\sigma$0.5
& 0.924
& 0.823
& 0.407
& 0.514
& 0.687
& 0.255\\
+gaussian $\sigma$0.7
& 0.927
& 0.846
& 0.437
& 0.575
& 0.721
& 0.245\\
+JPEG 75
& 0.969
& 0.896
& 0.477
& 0.670
& 0.715
& 0.164\\
+TV $\lambda$0.04
& 0.960
& 0.882
& 0.454
& 0.587
& 0.749
& 0.191\\
\bottomrule
\end{tabular}
\vspace{1mm}
\caption{Robustness to Image Preprocessing. }
\label{tab:img_preprocessing}
\end{table}

\noindent\textbf{Black-box Transfer across Models and Limitations.}

To evaluate the black-box transferability of our method, we conduct cross-model evaluations between Matrix-Game-2.0 \cite{matrixgame2} and Astra \cite{Astra}. Given their different default input resolutions, we resize the generated adversarial context images to match the target model's configuration prior to inference. As shown in Tab. \ref{tab:blackbox_transfer_models}: (1) Trained on Matrix-Game, test on Astra (Left): Our method demonstrates clear transferability. The video quality metrics notably decline compared to the clean baseline. Specifically, Background Consistency drops from 0.878 to 0.846, and Imaging Quality decreases from 0.553 to 0.515. This confirms the attack's effectiveness on an unseen architecture. (2) Trained on Astra, Test on Matrix-Game (Right): The adversarial transferability is severely limited, showing negligible performance degradation. This asymmetry highlights a key limitation: excessive resizing required to bridge resolution gaps likely disrupts the delicate pixel-level perturbations, thereby degrading transferability in black-box scenarios.

\begin{table}[H]
\centering
\small
\setlength{\tabcolsep}{4pt}
\renewcommand{\arraystretch}{1.18}

\begin{tabular*}{\textwidth}{@{\extracolsep{\fill}}lccc!{\hspace{6pt}\color{black!55}\vrule width 0.5pt\hspace{6pt}}ccc@{}}
\toprule
& \multicolumn{3}{c}{\textbf{Matrix-Game $\rightarrow$ Astra}}
& \multicolumn{3}{c}{\textbf{Astra $\rightarrow$ Matrix-Game}} \\
\cmidrule(lr){2-4} \cmidrule(lr){5-7}
\textbf{Method}
& \textbf{Back. Consis.}
& \textbf{Aesthetic}
& \textbf{Imaging Qual.}
& \textbf{Back. Consis}
& \textbf{Aesthetic}
& \textbf{Imaging Qual.} \\
\midrule
Clean & 0.878 & 0.492 & 0.553 & 0.955 &  0.488 &  0.659\\
Velocity-Min & 0.846 & 0.466 & 0.515 & 0.954 &  0.485 &  0.645\\
\bottomrule
\end{tabular*}
\vspace{1mm}
\caption{Black-box Transfer across Models.}
\label{tab:blackbox_transfer_models}
\end{table}
\newpage
\section{Implementation Details}
\label{app:implementation}
\subsection{Trajectory Sampling}
\subsubsection{Camera Trajectory Formulation}
\label{app:trajectory_parameterization}
To ensure the adversarial search space is both expressive and tractable, we represent the camera control sequence using a compact, low-dimensional parameterization. At any target frame $t$, the control signal is defined as a triplet $c_t = (\psi_t, f_t, s_t)$, representing the yaw angle, forward displacement, and lateral shift, respectively. This triplet is uniquely mapped to a relative camera pose matrix 
$P_t \in \mathbb{R}^{3 \times 4}$:$$ P_t = \begin{bmatrix} \cos \psi_t & 0 & \sin \psi_t & s_t \\ 0 & 1 & 0 & 0 \\ -\sin \psi_t & 0 & \cos \psi_t & -f_t \end{bmatrix} $$
The first $3 \times 3$ block encodes the yaw rotation around the vertical axis, while the translation vector captures horizontal and forward motions. For an autoregressive chunk comprising $T=8$ frames, the complete continuous trajectory is vectorized as $\tau = [\psi_1, f_1, s_1, \dots, \psi_T, f_T, s_T]^\top \in \mathbb{R}^{3T}$.To ensure the optimized trajectories remain physically plausible and strictly within the model's in-distribution control space, we enforce bound constraints $\mathcal{B}$ on both the absolute magnitudes and the temporal derivatives of the sequence. Specifically, the frame-wise controls are bounded by $\psi_{\max}=0.05$, $f_{\max}=0.05$, and $s_{\max}=0.025$. To prevent erratic, discontinuous camera jumps, the step-wise variations are strictly bounded by $\Delta_\psi=0.03$, $\Delta_f=0.03$, and $\Delta_s=0.015$. All sampled or optimized trajectories are projected onto this feasible convex set.

\subsubsection{Stochastic Trajectory Sampling via Random Walk}
\label{app:random_walk_trajectory}
For the baseline self-supervised objectives (Sec.\ref{objs}) where adaptive mining is not employed, we approximate the expectation over control signals through stochastic trajectory sampling. At each step of the Projected Gradient Descent (PGD) \cite{pgd}, a fresh trajectory is generated to prevent the adversarial perturbation from overfitting to a static camera motion. The initial frame control $c_1$ is drawn uniformly from the frame-wise feasible range. Subsequent frames follow a Gaussian random walk to ensure temporal coherence:$$ c_t = c_{t-1} + \eta_t, \quad \eta_t \sim \mathcal{N}(0, \Sigma_{rw}) $$where $\Sigma_{rw} = \operatorname{diag}(\sigma_\psi^2, \sigma_f^2, \sigma_s^2)$.
The resulting trajectory is subsequently clipped to satisfy the variation and range constraints defined in Sec.~\ref{app:trajectory_parameterization}. This unbiased stochastic exploration ensures the generated perturbation maintains generalizability across a wide distribution of camera motions.

\subsubsection{Trajectory-Adaptive Bi-Level Execution}\label{app:bilevel_trajectory_details} 
For the trajectory-adaptive bi-level attack (Sec. \ref{bi-attack}),
during the outer optimization loop, we search for trajectories that maximize the empirical adversarial loss. To account for the stochasticity of the denoising process, the fitness of a candidate trajectory $\tau$ is evaluated via Monte Carlo approximation over $M$ evaluation contexts (fixing the history length, timestep, and latent noise):$$ F(\tau;\delta) = \frac{1}{M} \sum_{j=1}^{M} \mathcal{L}_{atk}(x+\delta;\tau,\xi_j) $$For efficiency, we default to $M=1$. We maintain separate hard trajectory pools $\mathcal{P}_{\mathrm{hard}}$ tailored for different autoregressive history lengths, recognizing that the sensitivity of the model to specific camera motions diverges as the rollout progresses. During the inner loop, the PGD update dynamically queries these pools. For standard short-horizon steps, the attack defaults to the random-walk sampler to maintain broad robustness. For extended horizons where error accumulation is critical, the attack samples directly from $\mathcal{P}_{\mathrm{hard}}$, forcing the inner minimization to prioritize control sequences under which the current perturbation is least effective.

\subsubsection{Covariance Matrix Adaptation Evolution Strategy (CMA-ES)\cite{cma}}\label{app:cma_details}
\begin{figure}[H]
    \centering
    \includegraphics[width=1.0\linewidth]{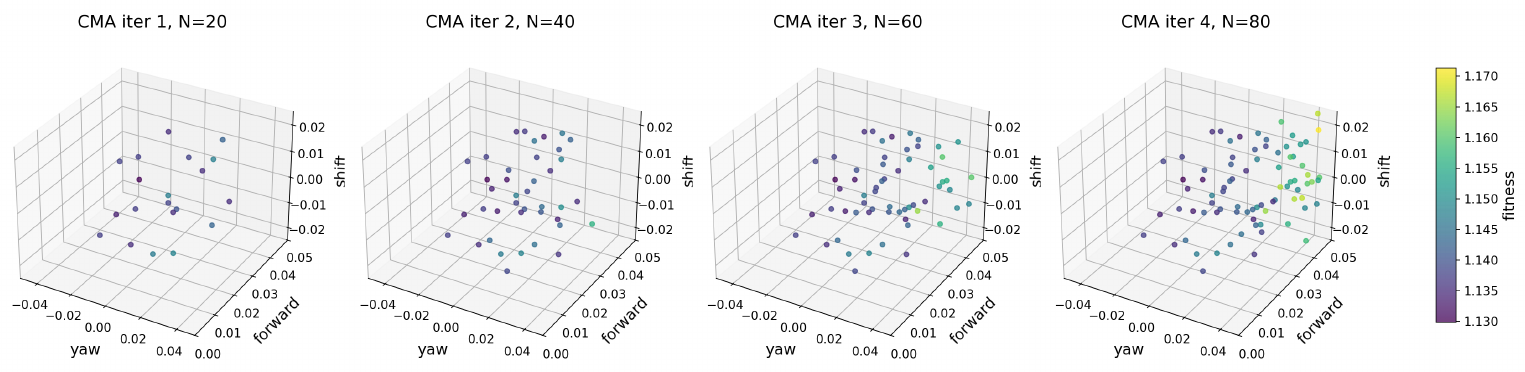}
    \caption{CMA trajectory updates.}
    \label{fig:cma update}
\end{figure}

The outer maximization over $\tau \in \mathbb{R}^{3T}$ presents a challenging non-convex optimization problem over a continuous sequence. Importantly, computing exact gradients through the multi-step autoregressive generation process is computationally prohibitive and prone to gradient shattering. We thus employ CMA-ES\cite{cma}, a derivative-free evolutionary algorithm, to efficiently navigate this space. At generation $g$, CMA-ES maintains a multivariate Gaussian search distribution parameterized by a mean trajectory $m^{(g)}$, a global step size $\sigma^{(g)}$, and a covariance matrix $C^{(g)}$:$$ q^{(g)}(\tau) = \mathcal{N}\left(m^{(g)}, (\sigma^{(g)})^2 C^{(g)}\right) $$A population of $\lambda$ candidate trajectories is sampled, projected into the feasible region, and evaluated against the fitness function $-F(\tau;\delta)$. Let $\{\tau^{(g)}_{i:\lambda}\}_{i=1}^\mu$ denote the top $\mu$ candidates sorted by fitness. The distribution mean is updated toward the weighted average of these elite candidates:$$ m^{(g+1)} = \sum_{i=1}^{\mu} w_i \tau^{(g)}_{i:\lambda} $$where $w_i > 0$ are the recombination weights. A critical advantage of CMA-ES for sequence mining is the adaptation of the covariance matrix $C^{(g)}$, which captures the strong temporal correlations inherent in hard camera motions (e.g., a coordinated continuous pan and forward zoom). The covariance is updated by integrating both the current elite population and an accumulated evolution path 
\begin{equation}
    \qquad
    C^{(g+1)}
    =
    (1-c_{\mathrm{cov}})C^{(g)}
    +
    c_{\mathrm{cov}}
    \sum_{i=1}^{\mu}
    w_i
    y_{i:\lambda}^{(g)}
    \left(y_{i:\lambda}^{(g)}\right)^\top
    \label{eq:cma_update}
\end{equation}
where $y_{i:\lambda}^{(g)} = (\tau_{i:\lambda}^{(g)}-m^{(g)})/\sigma^{(g)}$ is the normalized step. 
As shown in Fig.\ref{fig:cma update}, the trajectories are successively updated to higher velocity norm.
Upon termination of the CMA-ES generations, the globally top-ranked candidates across the evaluation history are distilled into the hard trajectory pool $\mathcal{P}_{\mathrm{hard}}$ for the inner PGD optimization.

\subsection{Additional Experimental Details}
\label{imple_details}
\noindent\textbf{Training.}
As discussed in Sec.\ref{label-free design}, the context frame available to users does not come with paired ground-truth videos or viewpoints annotations. To simulate realistic attack scenarios, we introduce two key adjustments: (i) we restrict training to the early denoising phase (timesteps ranging from 950 to 1000 by default) where target frames can be approximated as pure noise, and (ii) we simulate historical context by duplicating the context frame.

For conditioning, Astra uses 30 prompt variants generated via GPT-4o-mini, each semantically aligned with the context frames; we randomly sample one prompt per attack and a unique camera pose per latent frame.  Matrix-Game 2.0 requires no text prompts, so we randomly sample discrete actions per frame.

For each model, all four attack objectives are evaluated under identical training schedules (Tab.  \ref{train_para}). We conduct training on an A800 80GB GPU.

\vspace{-3mm}
\begin{table}[H]
\centering
\small
\begin{tabular}{lccccc}
\hline
\textbf{Model} & \textbf{VAE Ratio} & \textbf{Frames per Chunk}  & \textbf{Attack Budget} & \textbf{Step Size} & \textbf{Steps} \\
\hline
Astra & 4  & 8 & 0.05 & 0.005 & 300 \\
Matrix-Game 2.0 & 4  & 3 & 0.05 & 0.004 & 300 \\
\hline
\end{tabular}
\caption{Training hyperparameters for different models.}
\label{train_para}
\end{table}
\vspace{-5mm}

\noindent\textbf{Inference.}
We follow the default inference configurations (Tab. \ref{inf_para}) of the base models and generate videos autoregressively. For each context frame (input image), we produce 3 videos using different camera or action sequences to ensure result diversity.

\begin{table}[H]
\centering
\small
\renewcommand{\arraystretch}{1.2}
\resizebox{\linewidth}{!}{
\begin{tabular}{lccccc}
\hline
\textbf{Model} & \textbf{Frames per Chunk} & \textbf{Steps per Chunk} & \textbf{Target Latent Frames} & \textbf{Fps} & \textbf{Attn. Window}\\
\hline
Astra & 8 & 50 & 33 & 20 & 20 \\
Matrix-Game 2.0 & 3 & 3 & 60 & 5 & 4 \\
\hline
\end{tabular}
}
\caption{Inference settings for both models.}
\label{inf_para}
\end{table}

\noindent\textbf{Evaluations.}
While Astra is evaluated via standard VBench metrics, we adapt our approach for the longer-form Matrix-Game-2. Specifically, we employ VBench-Long for video quality evaluation and segment the original videos into shorter clips to precisely assess contextual consistency. This length-aware strategy ensures a more rigorous and reliable evaluation.
\newpage
\section{Extension Experiments on Matrix-Game-2 Variants}\label{sec:supp_extension_experiments}

\label{sec:supp_matrix_variants}
For the Matrix-Game-2 universal variants, we train the adversarial perturbation with an attack budget of $0.05$ and a step size of $0.002$ for a total of $700$ optimization steps. During inference, we follow the default Matrix-Game-2 configuration: the local attention window is set to $6$, each chunk contains $3$ latent frames, and generation is performed with $3$ sampling steps. For each input image, we generate $27$ latent frames in total.

\begin{figure}[H]
    \centering
    \includegraphics[width=1.0\linewidth]{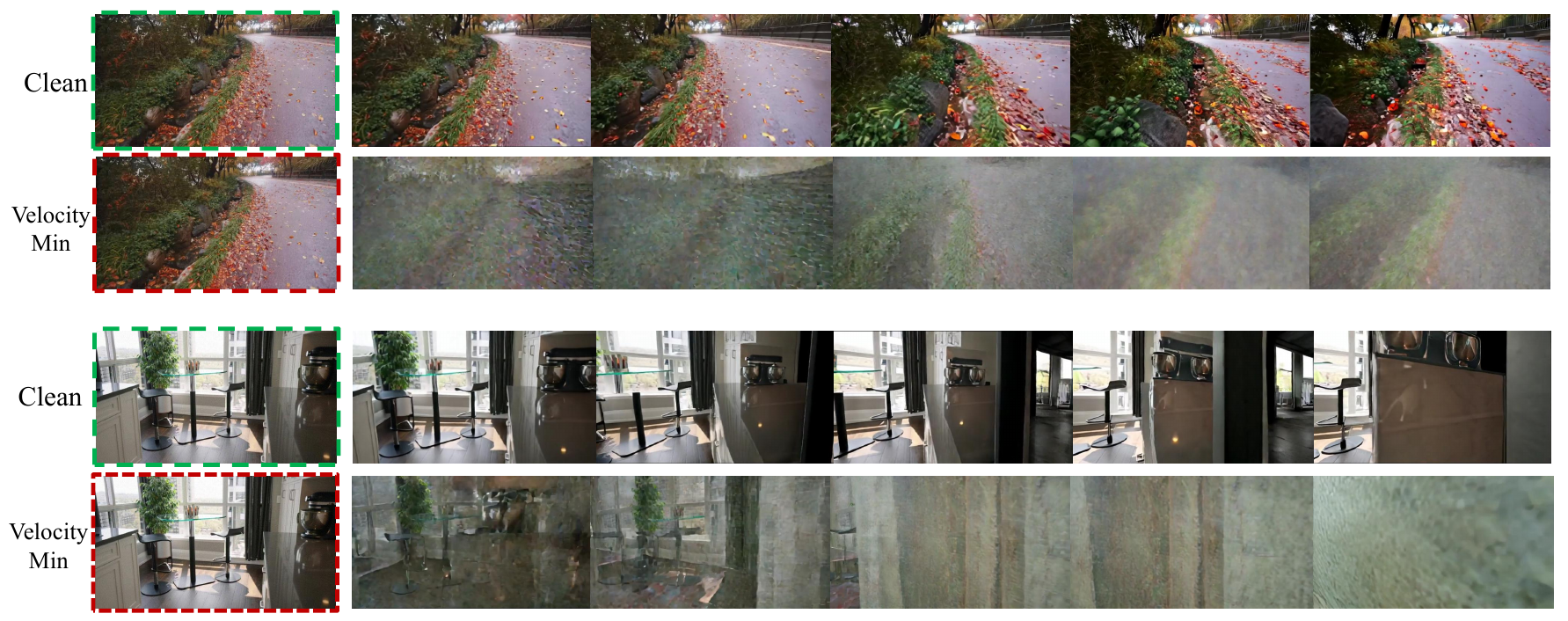}
    \vspace{-4mm}
    \caption{Attack Performance on Matrix-Game-2.0 (Universal).}
    \label{fig:matrix_uni_demo}
\end{figure}

As shown in Table \ref{tab:uni_results} and Figure \ref{fig:matrix_uni_demo}, \textsc{BadWorld} effectively generalizes to the Matrix-Game-2 Universal variant. The Velocity-Min objective triggers substantial degradation across all quantitative metrics, notably reducing Imaging Quality from 0.652 to 0.515 and Aesthetic Quality from 0.513 to 0.418. Qualitative results further confirm that the adversarial perturbation successfully induces structural collapse in the rollouts, validating the robustness of our framework across different model configurations.

\begin{table}[H]
\centering
\small
\resizebox{\textwidth}{!}{
\begin{tabular}{lcccccc}
\toprule
\textbf{Method} 
& \makecell{\textbf{I2V}\\ \textbf{Background} $\downarrow$}
& \makecell{\textbf{Background}\\ \textbf{Consistency} $\downarrow$}
& \makecell{\textbf{Aesthetic}\\ \textbf{Quality} $\downarrow$}
& \makecell{\textbf{Imaging}\\ \textbf{Quality} $\downarrow$}
& \textbf{CLIP-I$\downarrow$}
& \textbf{MEt3R$\uparrow$}\\
\midrule

Clean        & 0.961 & 0.932 & 0.513 & 0.652 & 0.755 & 0.162\\
Velocity-Min & 0.874 & 0.875 & 0.418 & 0.515 & 0.703 & 0.223\\

\bottomrule
\end{tabular}
}
\caption{Attack on Matrix-Game-2 (Universal)}
\label{tab:uni_results}
\end{table}

\newpage
\section{Additional Qualitative Results}
\label{sec:supp_qualitative_results}

This section provides additional qualitative results on Astra and Matrix-Game-2.0. 

\begin{figure}[H]
    \centering
    \includegraphics[width=1.0\linewidth]{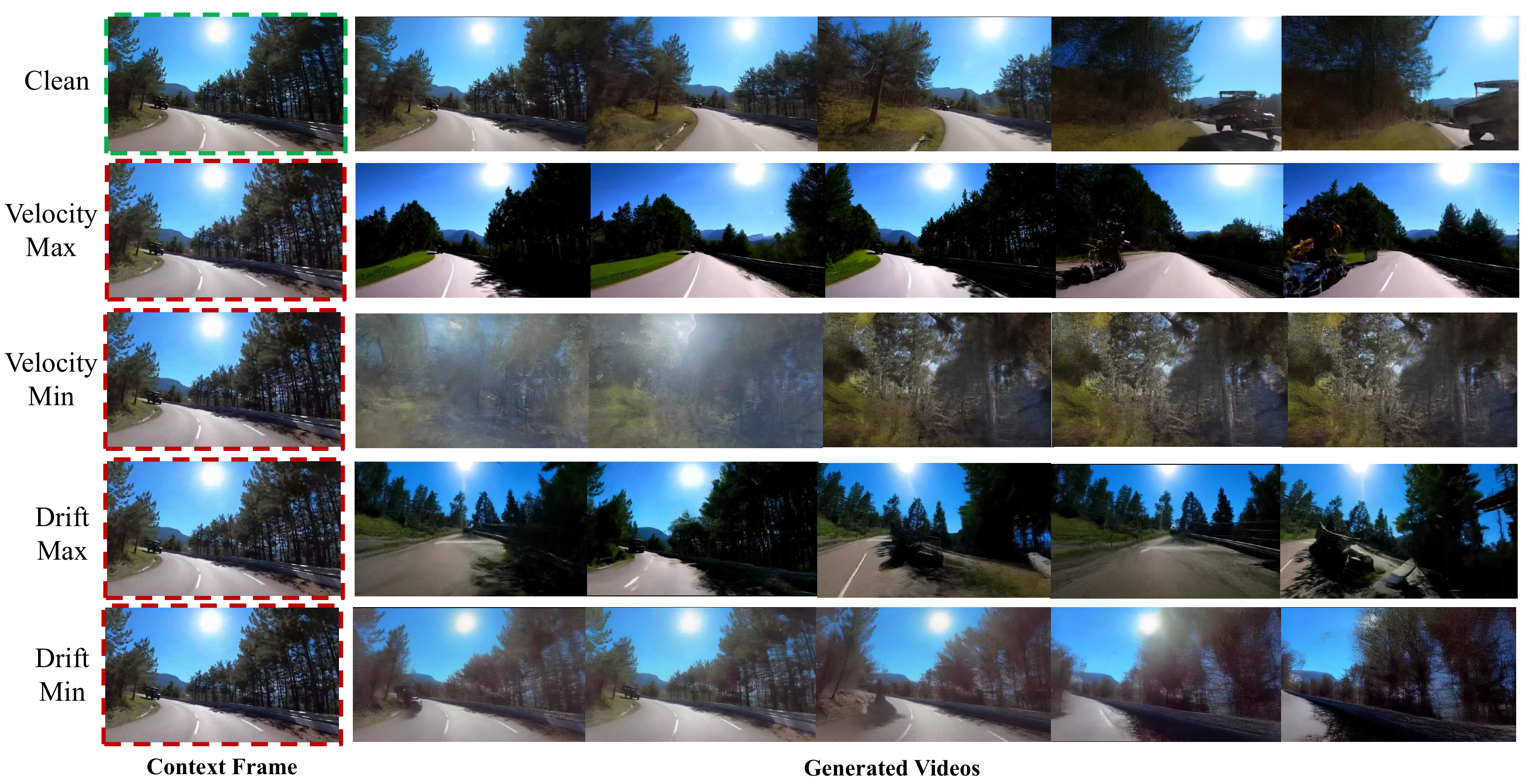}
    \caption{Qualitative Results on Astra.}
    \label{fig:astra_demo2}
\end{figure}
\vspace{5mm}
\begin{figure}[H]
    \centering
    \includegraphics[width=1.0\linewidth]{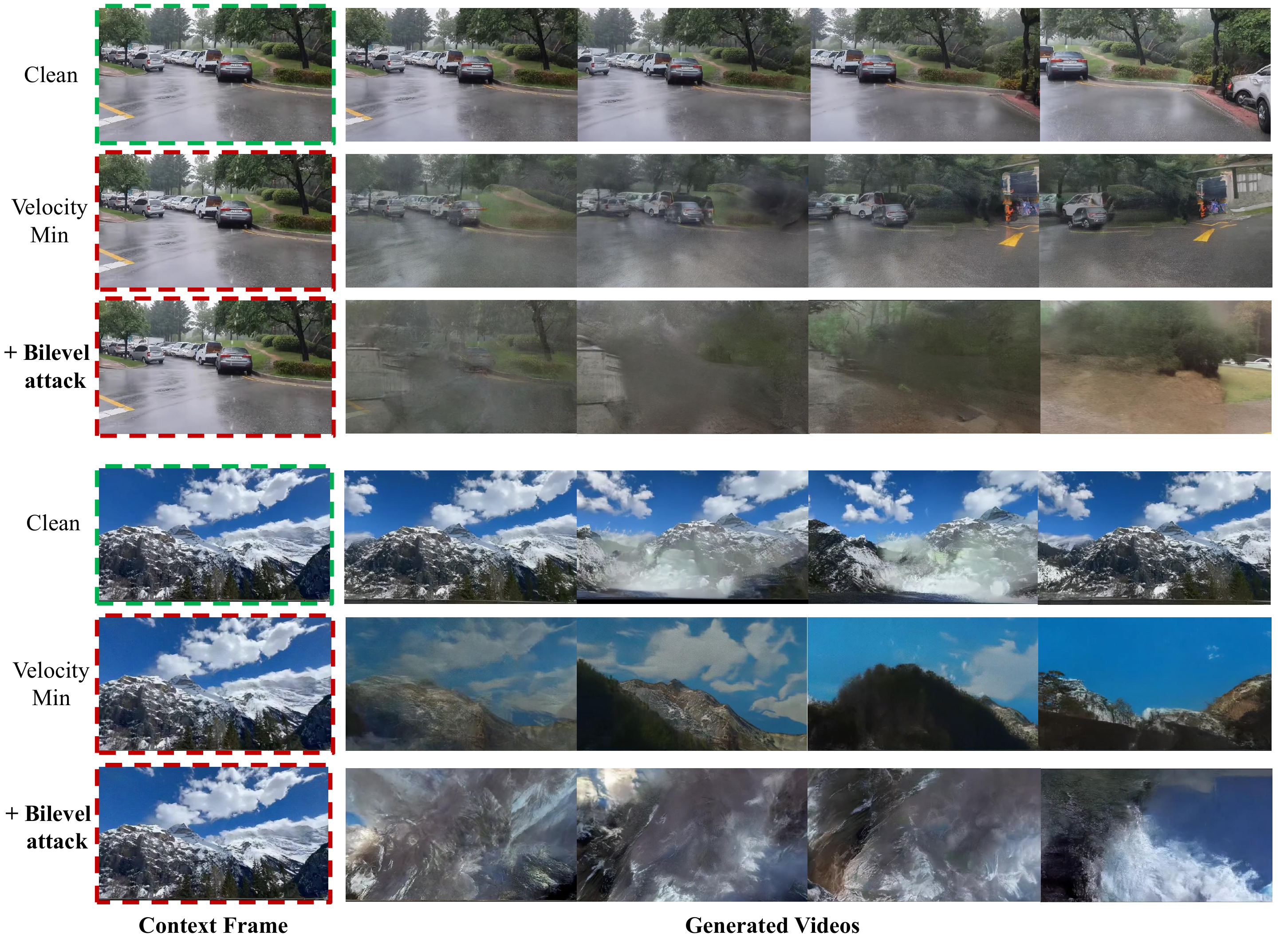}
    \caption{Performance of Bi-Level Attack.}
    \label{fig:bi_demo2}
\end{figure}
\newpage

\begin{figure}[H]
    \centering
    \includegraphics[width=0.9\linewidth]
    {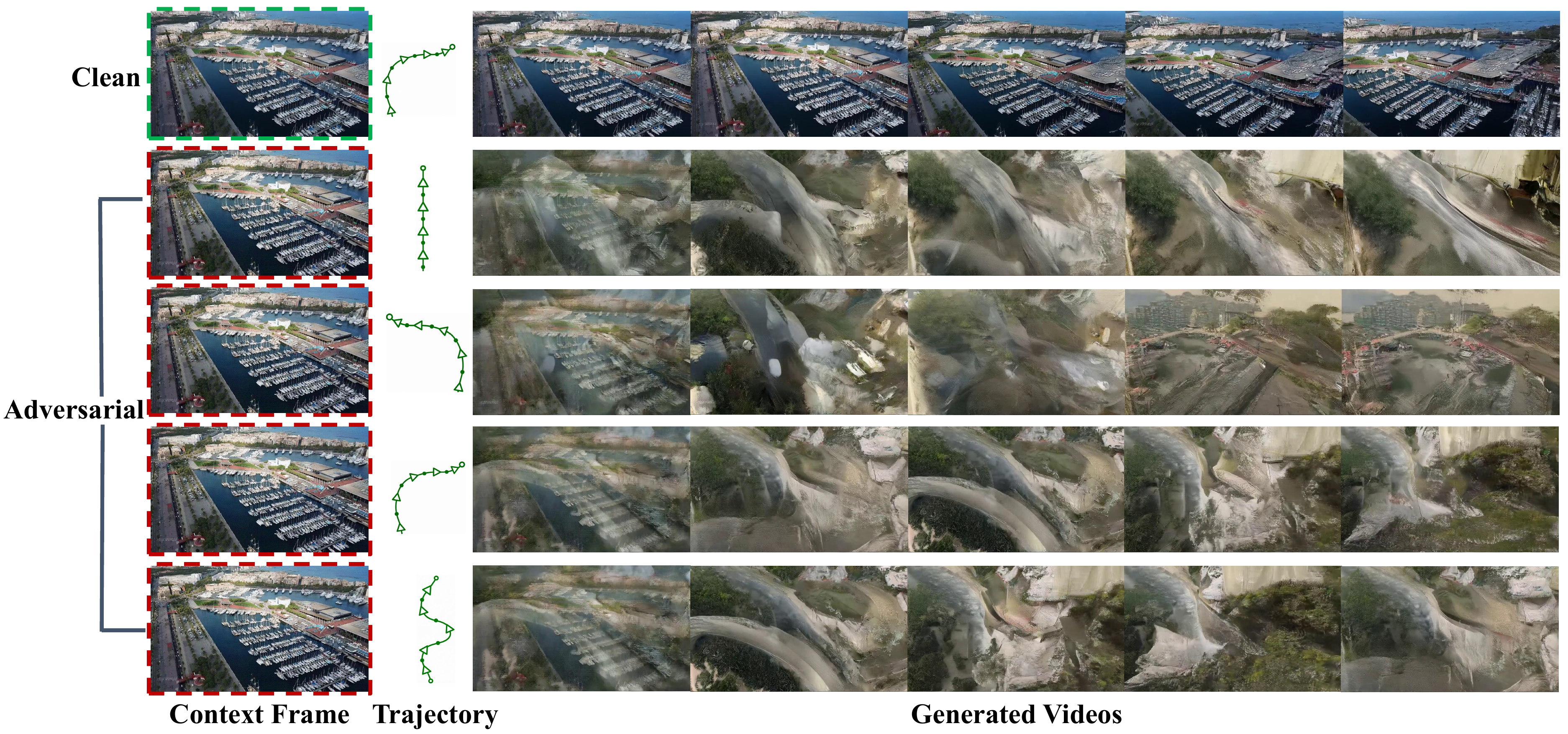}
    \vspace{-3mm}
    \caption{Attack performance unfer different camera trajectories.}
    \label{fig:diff_cam}
\end{figure}

\begin{figure}[H]
    \centering
    \includegraphics[width=0.9\linewidth]
    {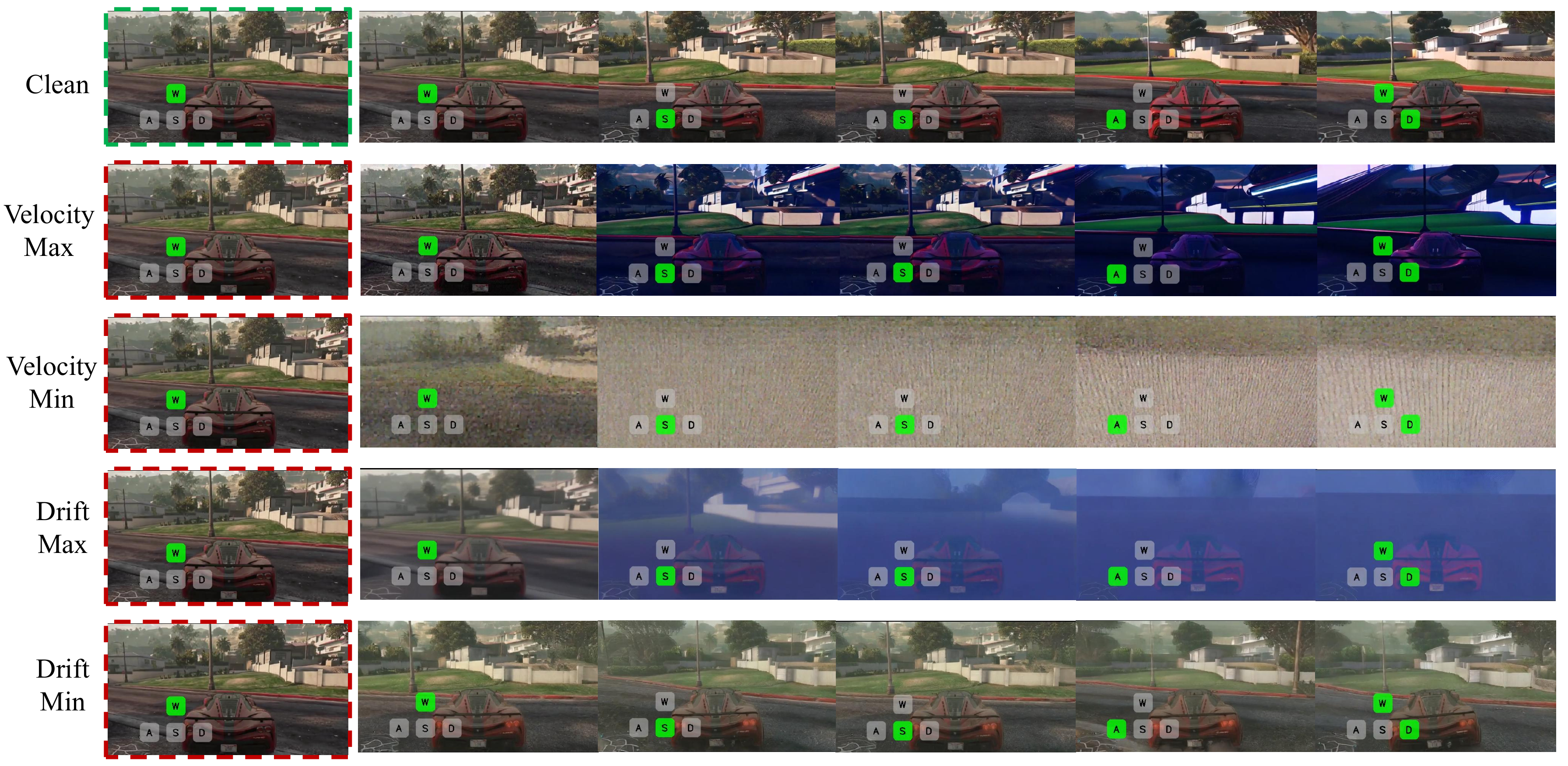}
    \vspace{-3mm}
    \caption{Qualitative Results on Matrix-Game-2.0.}
    \label{fig:matrix_demo2}
\end{figure}





\end{document}